\documentclass{article}

\usepackage{microtype}
\usepackage{graphicx}
\usepackage{subfigure}
\usepackage{booktabs} 
\usepackage{colortbl}
\usepackage[table]{xcolor}
\usepackage{hyperref}
\definecolor{lightgray}{gray}{0.92}
\definecolor{bestgreen}{RGB}{220,245,220}

\usepackage[accepted]{icml2025}

\usepackage{amsmath}
\usepackage{amssymb}
\usepackage{mathtools}
\usepackage{amsthm}

\usepackage[capitalize,noabbrev]{cleveref}

\theoremstyle{plain}

\theoremstyle{definition}

\theoremstyle{remark}

\usepackage[textsize=tiny]{todonotes}

\icmltitlerunning{3D Wavelet-Based Structural Priors}

\begin{document}

\twocolumn[
\icmltitle{3D Wavelet-Based Structural Priors for Controlled Diffusion in Whole-Body Low-Dose PET Denoising}



\icmlsetsymbol{equal}{$\dagger$}

\begin{icmlauthorlist}
\icmlauthor{Peiyuan Jing}{ZHAW,IC}
\icmlauthor{Yue Tang}{IC}
\icmlauthor{Chun-Wun Cheng}{Cam}
\icmlauthor{Zhenxuan Zhang}{IC}
\icmlauthor{Liutao Yang}{IC}
\icmlauthor{Thiago V. Lima}{Luc,ZHAW}
\icmlauthor{Klaus Strobel}{Luc}
\icmlauthor{Antoine Leimgruber}{Luc}
\icmlauthor{Angelica Aviles-Rivero}{Tsu}
\icmlauthor{Guang Yang}{equal,IC,nhl,Crc,Kcl}
\icmlauthor{Javier A. Montoya-Zegarra}{equal,ZHAW,Luc}

\end{icmlauthorlist}

\icmlaffiliation{ZHAW}{School of Engineering, Zurich University of Applied Sciences, CH}
\icmlaffiliation{IC}{Bioengineering Department and Imperial-X, Imperial College London, UK}
\icmlaffiliation{Cam}{DAMTP, University of Cambridge, UK}
\icmlaffiliation{Luc}{Lucerne University Teaching and Research Hospital, CH}
\icmlaffiliation{nhl}{National Heart and Lung Institute, Imperial College London, UK}
\icmlaffiliation{Crc}{Cardiovascular Research Centre, Royal Brompton Hospital, UK}
\icmlaffiliation{Kcl}{School of Biomedical Engineering \& Imaging Sciences, King's College London, UK}
\icmlaffiliation{Tsu}{Yau Mathematical Sciences Center, Tsinghua University, CN}

\icmlcorrespondingauthor{Javier Montoya}{javier.montoya@zhaw.ch}


\vskip 0.3in]



\printAffiliationsAndNotice{$\dagger$ co last authors} 

\begin{abstract}
Low-dose Positron Emission Tomography (PET) imaging reduces patient radiation exposure but suffers from increased noise that degrades image quality and diagnostic reliability. Although diffusion models have demonstrated strong denoising capability, their stochastic nature makes it challenging to enforce anatomically consistent structures, particularly in low signal-to-noise regimes and volumetric whole-body imaging. We propose Wavelet-Conditioned ControlNet (WCC-Net), a fully 3D diffusion-based framework that introduces explicit frequency-domain structural priors via wavelet representations to guide volumetric PET denoising. By injecting wavelet-based structural guidance into a frozen pretrained diffusion backbone through a lightweight control branch, WCC-Net decouples anatomical structure from noise while preserving generative expressiveness and 3D structural continuity. Extensive experiments demonstrate that WCC-Net consistently outperforms CNN-, GAN-, and diffusion-based baselines. On the internal $1/20$-dose test set, WCC-Net improves PSNR by $+1.21$~dB and SSIM by $+0.008$ over a strong diffusion baseline, while reducing structural distortion (GMSD) and intensity error (NMAE). Moreover, WCC-Net generalizes robustly to unseen dose levels ($1/50$ and $1/4$), achieving superior quantitative performance and improved volumetric anatomical consistency. 

\end{abstract}

\section{Introduction}
\label{sec:intro}

Positron emission tomography (PET) is a non-invasive molecular imaging modality widely used in clinical practice for the diagnosis and monitoring of various diseases, including cancer, cardiovascular disorders, and neurological conditions, owing to its high specificity and quantitative accuracy~\cite{berger2003positron,fletcher2008recommendations,granov2013positron}. However, a fundamental limitation of clinical PET lies in the trade-off between image quality and patient radiation exposure~\cite{granov2013positron}. High-quality PET imaging requires the administration of a sufficient amount of radiotracer, which inevitably increases the radiation dose to the patient~\cite{karp1991performance}. To reduce radiation exposure, clinical protocols often employ lower tracer doses, resulting in low-dose PET images that suffer from increased noise and reduced contrast, thereby degrading diagnostic reliability~\cite{miglioretti2013use}. Consequently, synthesizing normal-dose PET images from low-dose acquisitions while preserving critical diagnostic information is of significant clinical importance.

In recent years, deep learning has emerged as a dominant paradigm for PET image denoising, driven by advances in computational resources and the availability of large-scale medical imaging datasets~\cite{gong2018pet,cui2019pet}. Early approaches primarily relied on Convolutional Neural Networks (CNNs), which demonstrated promising performance in suppressing noise while preserving anatomical structures~\cite{zhou2021mdpet}. Subsequently, Generative Adversarial Network (GANs)-based methods introduced adversarial loss functions to enhance perceptual realism, producing images with improved visual quality~\cite{fu2023aigan}. More recently, Diffusion Model (DMs)–based approaches~\cite{ho2020denoising} have gained increasing attention, and various diffusion-based variants have been explored for PET denoising~\cite{jiang2023pet}, due to their strong generative ability and robustness in modeling complex data distributions.

However, despite achieving promising performance, existing DM–based denoising methods remain limited by their distribution-based nature, which often weakens their ability to effectively exploit explicit structural guidance and spatial-domain information~\cite{jiang2023pet,shen2023pet}. PET images are sensitive to noise in high-frequency components~\cite{miglioretti2013use}; consequently, unconditioned or weakly conditioned DMs face an inherent structure–detail trade-off: suppressing pervasive high-frequency noise while preserving fine anatomical edges~\cite{yu2025robust}. When conditioning is applied solely on raw low-dose PET images in the spatial domain, structural information is entangled with noise, introducing ambiguity in the guidance signal. As a result, DMs tend to favor overall perceptual consistency, often at the expense of fine anatomical details, leading to blurred edges or missing structures that are critical for accurate clinical diagnosis~\cite{yu2025robust,jiang2023pet}.

The Discrete Wavelet Transform (DWT)~\cite{heil1989continuous} provides a powerful mechanism for multi-dimensional signal analysis by decomposing PET images into frequency- and scale-specific bands, enabling explicit separation of robust low-frequency structural information from noise-dominated high-frequency details~\cite{tian2023multi}. Meanwhile, ControlNet~\cite{zhang2023adding} offers an effective means of injecting structured external guidance into DMs without disrupting their pretrained generative capacity. Leveraging the complementary strengths of these two components, we propose Wavelet-Conditioned ControlNet (WCC-Net), a novel architecture that provides explicit and noise-robust structural guidance to a 3D Denoising Diffusion Probabilistic Model (DDPM) for whole-body low-dose PET denoising. By exploiting the stability of frequency-domain representations, WCC-Net mitigates the structural degradation commonly observed in spatially conditioned DMs: low-frequency wavelet coefficients guide volumetric structural preservation, while stochastic noise suppression is delegated to the frozen diffusion backbone, thereby preserving 3D anatomical continuity. Our main contributions are threefold:
\begin{itemize}
    \item We propose a fully 3D wavelet-conditioned diffusion framework for whole-body PET denoising that preserves volumetric anatomical continuity.
    \item We introduce a decoupled wavelet-based conditioning strategy that injects frequency-domain structural priors via a ControlNet-style trainable branch into a frozen diffusion backbone.
    \item Extensive validation on ultra-low-dose whole-body PET, demonstrating consistent improvements over multiple baselines across seen and unseen dose levels, with improved structural fidelity and quantitative accuracy.
\end{itemize}

\section{Related Work}

% ...

% ...


\subsection{Low-dose PET Image Denoising}\label{subsec:pet_denoising}

Early approaches to low-dose PET image denoising relied on traditional model-driven techniques, such as non-local means filtering and block-matching-based methods~\cite{nlm,bm3d}. While effective in suppressing noise, these methods often oversmooth images and struggle to preserve fine anatomical details and quantitative accuracy under complex noise conditions~\cite{fu2023aigan}. With the advent of deep learning, data-driven methods have become the dominant paradigm for PET denoising. CNN-based architectures, such as 3D U-Net, REDCNN, and EDCNN~\cite{3dunet,redcnn,edcnn}, have demonstrated improved noise reduction and structural preservation by learning hierarchical spatial features. GAN-based methods further enhance perceptual quality through adversarial learning, but may introduce locally inconsistent textures or hallucinated details~\cite{wang20183D-cGAN}. More recently, diffusion-based models have shown strong denoising capability due to their powerful generative priors~\cite{yu2025robust}. Despite these advances, most existing deep learning approaches rely on spatial-domain conditioning, where anatomical structure remains entangled with noise, particularly in ultra-low-dose settings.

\subsection{Discrete Wavelet Transform in Image Denoising}\label{subsec:dwt_denoising}

Incorporating explicit model-based priors into deep learning models has proven effective for image denoising~\cite{tian2023multi}. Among such priors, the Discrete Wavelet Transform (DWT) is particularly attractive due to its ability to provide multi-scale frequency-domain representations that separate stable structural content from noise~\cite{heil1989continuous}. Accordingly, wavelet transforms have been integrated into deep learning models to complement data-driven learning with frequency-aware inductive biases~\cite{tian2023multi}. More recently, Lyu \emph{et al.}~\cite{lyu2025wid} incorporated wavelet representations into PET denoising by coupling wavelet decomposition with the diffusion process. In contrast, we propose Wavelet-Conditioned ControlNet (WCC-Net), which leverages wavelet-derived frequency information purely as a conditioning prior. By injecting wavelet-based structural cues through a dedicated control branch while preserving a frozen spatial-domain diffusion backbone, WCC-Net decouples structural guidance from the generative process and enables controllable, anatomy-aware PET denoising.

\subsection{ControlNet in Diffusion Models}\label{subsec:controlnet}
Recent advances in diffusion modeling aim to enable fine-grained control through auxiliary conditioning signals while preserving the expressive power of large pretrained backbones. In this context, Zhang \emph{et al.}~\cite{zhang2023adding} proposed ControlNet, a conditional control framework that augments DMs with parallel, trainable branches while keeping the original backbone frozen. Formally, let $\mathcal{F}_{\theta}(\cdot)$ denote a pretrained network block with parameters $\theta$ within a DM. ControlNet introduces a trainable copy $\mathcal{F}_{\phi}(\cdot)$ to process an auxiliary conditioning signal $\mathbf{c}$, whose features are injected into the frozen backbone via zero-initialized convolutions:
\begin{equation}
\tilde{\mathbf{y}}
=
\mathcal{F}_{\theta}(\mathbf{x})
+
\mathcal{Z}\!\left(
\mathcal{F}_{\phi}(\mathbf{x} + \mathcal{Z}(\mathbf{c}))
\right),
\label{eq:controlnet}
\end{equation}
where $\mathcal{Z}(\cdot)$ denotes a zero-initialized convolution. This design strategy gradually introduces the conditional pathway during optimization, while preserving the pretrained DM at initialization. By injecting conditioning information at multiple resolutions through residual connections, ControlNet enables structured and spatially consistent generation without compromising the underlying generative prior~\cite{yu2025adaptive,zhang2023adding}. 
However, most prior work focuses on spatial-domain conditioning, whereas the choice of conditioning representation becomes critical in low signal-to-noise regimes such as PET denoising.

\section{Method}

To provide noise-robust structural guidance while preserving the generative capacity of pretrained diffusion models, we propose Wavelet-Conditioned ControlNet (WCC-Net), as illustrated in Fig.~\ref{fig:WCC}. WCC-Net instantiates the conditioning signal $\mathbf{c}$ as an explicit frequency-domain structural prior derived from low-dose PET images. Specifically, Wavelet-domain representations are injected through a trainable control branch, serving as a principled inductive bias to decouple structural information from noise and to enforce anatomical consistency.
We begin by reviewing the diffusion backbone
(Subsec.~\ref{subsec:diffusion_backbone}). Next, we introduce the proposed Wavelet-based conditioning prior and its integration
(Subsec.~\ref{subsec:wavelet_prior}).

\subsection{Diffusion Backbone}\label{subsec:diffusion_backbone}

\begin{figure*}[h!t!]
\centering
\includegraphics[width=0.8\linewidth]{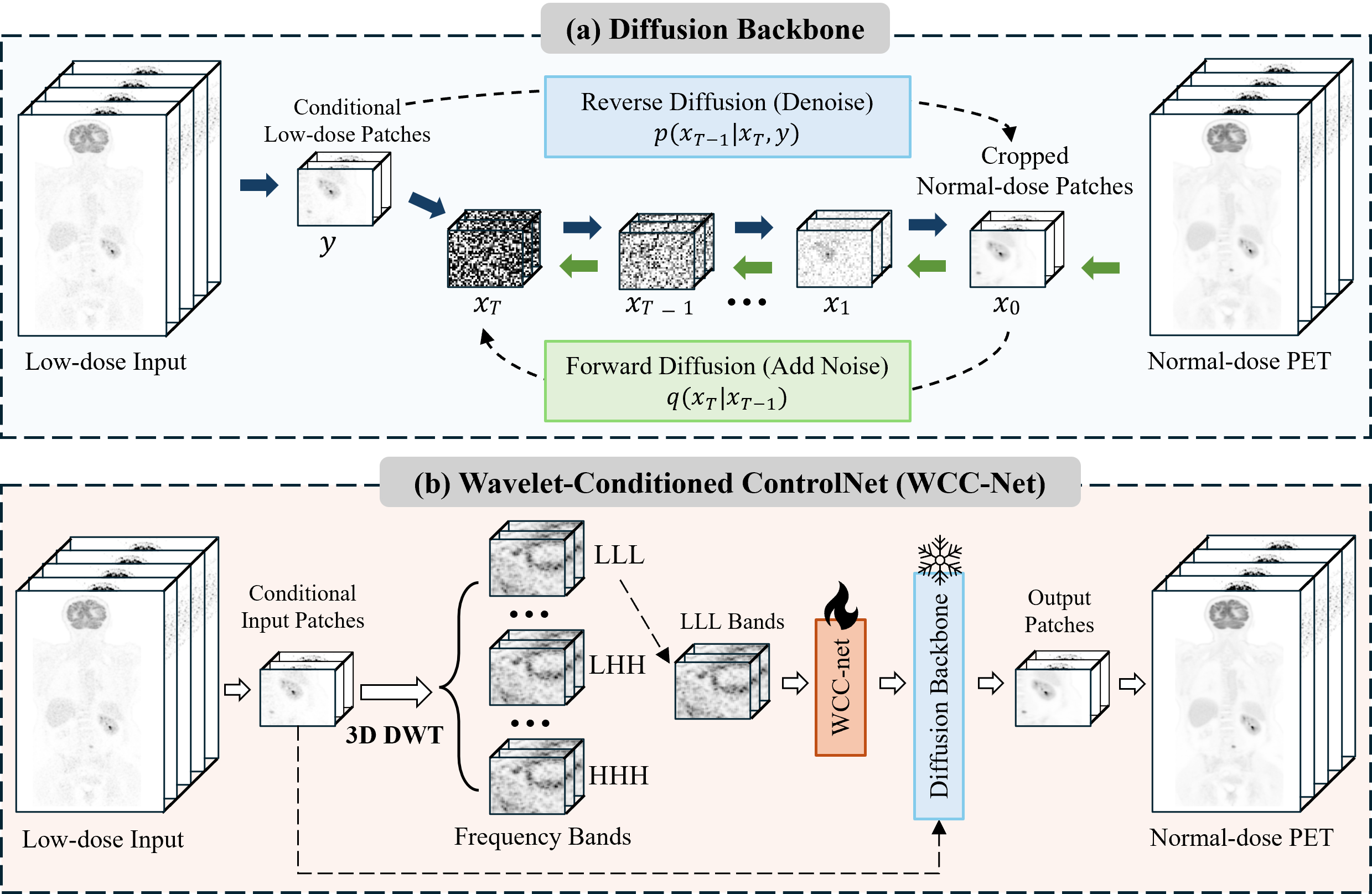}
\caption{
Overview of the proposed Wavelet-Conditioned ControlNet (WCC-Net).
(a) Diffusion backbone for low-dose to normal-dose PET patch denoising. (b) 
WCC-Net extracts multi-scale frequency structural priors via a 3D Discrete Wavelet Transform and injects them as conditional guidance into a frozen diffusion backbone, enabling fine-grained structure-aware and anatomically consistent PET denoising.
}
\label{fig:WCC}
\end{figure*}

As shown in Fig.~\ref{fig:WCC}(a), we adopt a conditional Denoising Diffusion Probabilistic Model (DDPM) as the backbone for low-dose to normal-dose PET image denoising. Let $x_0 \in \mathbb{R}^{H \times W \times D}$ denote a normal-dose PET image patch that belongs to the real normal-dose data distribution $q$, and let $y \in \mathbb{R}^{H \times W \times D}$ denote the corresponding low-dose PET patch used as conditional input. The forward diffusion process gradually corrupts the clean normal-dose PET patch $x_0$ with Gaussian noise over $T$ time steps, forming a Markov chain defined as:
\begin{equation}
q(x_t \mid x_{t-1}) = \mathcal{N}\left(x_t; \sqrt{1-\beta_t} x_{t-1}, \beta_t \mathbf{I}\right),
\end{equation}
where $\{\beta_t\}_{t=1}^T$ is a predefined variance schedule. By composition, $x_t$ can be sampled directly from $x_0$ as:
\begin{equation}
q(x_t \mid x_0) = \mathcal{N}\left(x_t; \sqrt{\bar{\alpha}_t} x_0, (1-\bar{\alpha}_t)\mathbf{I}\right),
\end{equation}
with $\alpha_t = 1 - \beta_t$ and $\bar{\alpha}_t = \prod_{s=1}^t \alpha_s$.

The reverse denoising process aims to recover $x_0$ from pure noise $x_T \sim \mathcal{N}(0, \mathbf{I})$ by learning a parameterized reverse transition $p_{\theta}$, which is defined as
\begin{equation}
p_\theta(x_{t-1} \mid x_t, y) = \mathcal{N}\left(x_{t-1}; \mu_\theta(x_t, t, y), \Sigma_\theta(x_t, t, y)\right),
\end{equation}
where $\mu_\theta$ and $\Sigma_\theta$ are predicted by a 3D U-Net conditioned on the low-dose PET image $y$. The model is trained to predict the injected noise $\epsilon$ using the objective
\begin{equation}
\mathcal{L}_{\text{DDPM}} = \mathbb{E}_{x_0, y, \epsilon, t}
\left[
\left\| \epsilon - \epsilon_\theta(x_t, t, y) \right\|_2^2
\right].
\end{equation}

This conditional diffusion backbone serves as a powerful generative prior capable of synthesizing realistic normal-dose PET images. However, when conditioned solely on noisy spatial-domain inputs, the model may struggle to disentangle anatomical structure from noise, motivating the introduction of explicit structural guidance.
\begin{table*}[t]
\centering
\footnotesize
\setlength{\tabcolsep}{3pt}
\renewcommand{\arraystretch}{0.95}
\begin{tabular}{lcccc}
\toprule
\multicolumn{5}{c}{\textbf{Internal (1/20 Dose, Seen during training)}} \\
\midrule
\textbf{Method} 
& \textbf{PSNR (dB) $\uparrow$} 
& \textbf{SSIM $\uparrow$} 
& \textbf{GMSD $\downarrow$}
& \textbf{NMAE $\downarrow$}\\ 
\midrule 
Low-dose   
& 38.545 $\pm$ 1.798** 
& 0.948 $\pm$ 0.018** 
& 0.042 $\pm$ 0.012** 
& 0.199 $\pm$ 0.036**\\
\midrule
\rowcolor{lightgray}
BM3D~\cite{bm3d}       
& 41.068 $\pm$ 2.000** 
& 0.976 $\pm$ 0.010** 
& 0.020 $\pm$ 0.008** 
& 0.130 $\pm$ 0.024** \\
NLM~\cite{nlm}      
& 40.386 $\pm$ 1.535** 
& 0.968 $\pm$ 0.008** 
& 0.019 $\pm$ 0.005** 
& 0.161 $\pm$ 0.020** \\
\rowcolor{lightgray}
3D cGAN~\cite{wang20183D-cGAN}      
& 40.721 $\pm$ 1.293** 
& 0.977 $\pm$ 0.006** 
& 0.014 $\pm$ 0.003** 
& 0.127 $\pm$ 0.015** \\
3D UNet~\cite{3dunet}    
& 41.810 $\pm$ 1.377** 
& 0.972 $\pm$ 0.007** 
& 0.013 $\pm$ 0.002* 
& 0.120 $\pm$ 0.015** \\
\rowcolor{lightgray}
3D REDCNN~\cite{redcnn}  
& 42.185 $\pm$ 1.427** 
& 0.981 $\pm$ 0.005* 
& 0.012 $\pm$ 0.003* 
& 0.119 $\pm$ 0.016** \\
3D EDCNN~\cite{edcnn}   
& 42.355 $\pm$ 1.579** 
& 0.983 $\pm$ 0.006* 
& 0.013 $\pm$ 0.003* 
& 0.116 $\pm$ 0.017** \\
\rowcolor{lightgray}
3D DDPM~\cite{yu2025robust}    
& 42.483 $\pm$ 1.593** 
& 0.983 $\pm$ 0.005** 
& 0.012 $\pm$ 0.004** 
& 0.115 $\pm$ 0.016** \\
\rowcolor{bestgreen}
\textbf{WCC-Net} 
& \textbf{43.594 $\pm$ 1.404} 
& \textbf{0.984 $\pm$ 0.005} 
& \textbf{0.011 $\pm$ 0.003} 
& \textbf{0.111 $\pm$ 0.014} \\
\midrule

\multicolumn{5}{c}{\textbf{External (1/50 Dose, Unseen dose level)}} \\
\midrule
\textbf{Method} 
& \textbf{PSNR (dB) $\uparrow$} 
& \textbf{SSIM $\uparrow$} 
& \textbf{GMSD $\downarrow$}
& \textbf{NMAE $\downarrow$} \\
\midrule
Low-dose   
& 34.297 $\pm$ 2.025** 
& 0.894 $\pm$ 0.034** 
& 0.087 $\pm$ 0.022** 
& 0.324 $\pm$ 0.066** \\
\midrule
\rowcolor{lightgray}
BM3D~\cite{bm3d}       
& 36.358 $\pm$ 2.381** 
& 0.940 $\pm$ 0.026** 
& 0.055 $\pm$ 0.020** 
& 0.226 $\pm$ 0.054** \\
NLM~\cite{nlm}        
& 37.603 $\pm$ 2.278** 
& 0.951 $\pm$ 0.018** 
& 0.035 $\pm$ 0.016** 
& 0.190 $\pm$ 0.038** \\
\rowcolor{lightgray}
3D cGAN~\cite{wang20183D-cGAN}      
& 38.459 $\pm$ 1.694** 
& 0.963 $\pm$ 0.014** 
& 0.024 $\pm$ 0.008** 
& 0.173 $\pm$ 0.029** \\
3D UNet~\cite{3dunet}    
& 38.465 $\pm$ 1.756** 
& 0.958 $\pm$ 0.014** 
& 0.018 $\pm$ 0.005** 
& 0.163 $\pm$ 0.030** \\
\rowcolor{lightgray}
3D REDCNN~\cite{redcnn}  
& 39.715 $\pm$ 1.950** 
& 0.968 $\pm$ 0.013** 
& 0.018 $\pm$ 0.006** 
& 0.169 $\pm$ 0.040** \\
3D EDCNN~\cite{edcnn}   
& 39.975 $\pm$ 1.711** 
& 0.970 $\pm$ 0.011** 
& 0.017 $\pm$ 0.005* 
& 0.158 $\pm$ 0.026** \\
\rowcolor{lightgray}
3D DDPM~\cite{yu2025robust}     
& 39.515 $\pm$ 1.847** 
& 0.964 $\pm$ 0.010** 
& 0.019 $\pm$ 0.005** 
& 0.151 $\pm$ 0.031** \\
\rowcolor{bestgreen}
\textbf{WCC-Net} 
& \textbf{40.754 $\pm$ 1.803} 
& \textbf{0.976 $\pm$ 0.011} 
& \textbf{0.014 $\pm$ 0.006} 
& \textbf{0.132 $\pm$ 0.027} \\
\midrule

\multicolumn{5}{c}{\textbf{External (1/4 Dose, Unseen dose level)}} \\
\midrule
\textbf{Method} 
& \textbf{PSNR (dB) $\uparrow$} 
& \textbf{SSIM $\uparrow$} 
& \textbf{GMSD $\downarrow$}
& \textbf{NMAE $\downarrow$} \\
\midrule
Low-dose   
& 45.612 $\pm$ 1.827** 
& 0.989 $\pm$ 0.004* 
& 0.009 $\pm$ 0.003* 
& 0.083 $\pm$ 0.015* \\
\midrule
\rowcolor{lightgray}
BM3D~\cite{bm3d}        
& 44.347 $\pm$ 1.534** 
& 0.980 $\pm$ 0.002** 
& 0.009 $\pm$ 0.002* 
& 0.097 $\pm$ 0.011** \\
NLM~\cite{nlm}        
& 43.124 $\pm$ 1.139** 
& 0.973 $\pm$ 0.005** 
& 0.014 $\pm$ 0.003** 
& 0.120 $\pm$ 0.015** \\
\rowcolor{lightgray}
3D cGAN~\cite{wang20183D-cGAN}      
& 42.413 $\pm$ 1.316** 
& 0.985 $\pm$ 0.003** 
& 0.012 $\pm$ 0.002** 
& 0.098 $\pm$ 0.009** \\
3D UNet~\cite{3dunet}    
& 43.642 $\pm$ 1.205** 
& 0.979 $\pm$ 0.004** 
& 0.012 $\pm$ 0.002** 
& 0.095 $\pm$ 0.009** \\
\rowcolor{lightgray}
3D REDCNN~\cite{redcnn}  
& 43.561 $\pm$ 1.013** 
& 0.980 $\pm$ 0.004** 
& 0.011 $\pm$ 0.002** 
& 0.125 $\pm$ 0.013** \\
3D EDCNN~\cite{edcnn}   
& 44.209 $\pm$ 1.327** 
& 0.986 $\pm$ 0.003** 
& 0.010 $\pm$ 0.002* 
& 0.101 $\pm$ 0.009** \\
\rowcolor{lightgray}
3D DDPM~\cite{yu2025robust}     
& 44.784 $\pm$ 1.528** 
& 0.981 $\pm$ 0.003** 
& 0.010 $\pm$ 0.004* 
& 0.089 $\pm$ 0.010** \\
\rowcolor{bestgreen}
\textbf{WCC-Net} 
& \textbf{45.235 $\pm$ 1.351} 
& \textbf{0.992 $\pm$ 0.003} 
& \textbf{0.007 $\pm$ 0.003} 
& \textbf{0.084 $\pm$ 0.013} \\
\bottomrule
\end{tabular}
\caption{Quantitative comparison of PET image denoising performance across different dose levels.
* and ** denote statistically significant differences compared with WCC-Net at $p<0.05$ and $p<0.01$, respectively.}
\label{tab:quant_results}
\end{table*}
\subsection{Wavelet-Based Conditioning Prior}\label{subsec:wavelet_prior}


Given a low-dose PET image patch $y \in \mathbb{R}^{H \times W \times D}$, we apply a 3D Discrete Wavelet Transform (DWT) to decompose $y$ into multiple frequency subbands:
\begin{equation}
\mathcal{W}(y) = \{ y^{\text{LLL}}, y^{\text{LLH}}, y^{\text{LHL}}, \ldots, y^{\text{HHH}} \},
\end{equation}
\label{eq:discrete_wavelet_transform}
where each letter $\text{L}$ or $\text{H}$ denotes low- or high-pass filtering along one spatial dimension, respectively. 
Specifically, $y^{\text{LLL}}$ represents the low-frequency (coarse structural) component across all three dimensions, while subbands containing $\text{H}$ capture progressively higher-frequency details such as edges and noise along the corresponding axes.

In low-dose PET imaging, the low-frequency wavelet component ($y^{\text{LLL}}$) predominantly encodes global coarse anatomical structure together with low-frequency intensity distribution, reflecting regional uptake patterns. Subbands containing a single high-pass component (e.g., $y^{\text{LLH}}$, $y^{\text{LHL}}$) capture directional fine-grained details, while the highest-frequency subband ($y^{\text{HHH}}$) is dominated by high-frequency fluctuations that are largely attributable to noise~\cite{li2025back}.
In our work, we adopt the Haar wavelet due to its orthogonality, computational efficiency, and effectiveness in representing piecewise-smooth signals, making it well suited for capturing coarse anatomical structures in medical images~\cite{mallat2002theory}. We deliberately adopt a single-level wavelet decomposition, as deeper decompositions increase channel complexity and complicate alignment with ControlNet injection. Given our focus on structural anchoring rather than full spectral modeling, this design provides a principled and efficient trade-off.

We denote a generic Wavelet-based prior extracted from a subset of subbands as:
\begin{equation}
\mathbf{c}_{\text{wav}} = \mathcal{W}_{\mathcal{S}}(y),
\end{equation}\label{eq:wavelet_prior}
where $\mathcal{S}$ indexes the selected subbands. Unless stated otherwise, the default configuration emphasizes low-frequency subbands, while alternative instantiations using individual high-frequency subbands and combinations of remaining Wavelet components are evaluated in Sec.~\ref{sec:results}.

To integrate the Wavelet prior, we substitute the conditioning signal in Eq.~\eqref{eq:controlnet} as: $\mathbf{c} \leftarrow \mathbf{c}_{\text{wav}}$. The Wavelet prior ($\mathbf{c}_{\text{wav}}$) is first processed by a lightweight embedding module consisting of a 3D transposed convolution that aligns its spatial resolution and channel dimensionality with the $\mathbb{R}^{H \times W \times D}$ feature space of the diffusion backbone. The embedded Wavelet features serve as the conditioning input $c$ to the trainable conditional pathway. The diffusion backbone remains frozen, and WCC-Net is initialized as a trainable copy of the corresponding encoder blocks of the pretrained 3D diffusion U-Net.

To preserve the integrity of the pretrained generative prior, the extracted Wavelet-conditioned priors are injected into the diffusion backbone via zero-initialized $1 \times 1 \times 1$ convolution layers (ZeroConv) before being added to the corresponding skip connections. For the l-th resolution level, the modified skip feature is given by:
\begin{equation}
\tilde{\mathbf{h}}_l
\;=\;
\mathbf{h}_l
\;+\;
\underbrace{\mathrm{ZeroConv}_l\!\left( \mathbf{c}_{\text{wav}}\right)}_{\text{conditional injection}},
\end{equation}\label{eq:wavelet_pathway}
where $\mathbf{h}_l$ denotes the original skip feature of the frozen diffusion backbone at level $l$. Each $\mathrm{ZeroConv}_l(\cdot)$ is initialized with zero weights so that the conditional pathway is inactive at initialization, recovering the original pretrained diffusion model at the start of training. Thus, the influence of the Wavelet-based structural prior is introduced progressively in a residual and architecture-preserving manner.

During optimization, only the parameters $\phi$ of WCC-Net are updated, whilst the diffusion backbone parameters $\theta$ remain frozen.
The noise prediction network is thus augmented to incorporate Wavelet-based structural guidance, yielding the refined denoising objective:
\begin{equation}
\mathcal{L}_{\text{WCC}} =
\mathbb{E}_{x_0, y, \epsilon, t}
\left[
\left\| \epsilon -
\epsilon_{\theta,\phi}
\bigl(x_t, t, y, \mathbf{c}_{\text{wav}}\bigr)
\right\|_2^2
\right],
\end{equation}\label{eq:wavelet_denoising_loss}
where $\epsilon_{\theta,\phi}$ denotes the noise predictor conditioned on both the low-dose PET input and the embedded Wavelet-based prior.

By freezing the diffusion backbone and injecting Wavelet-domain information through WCC-Net, the proposed framework supports anatomically consistent structure while allowing the diffusion process to focus on stochastic noise removal. This separation introduces a frequency-domain inductive bias without compromising the expressive capacity of the pretrained backbone. The additional computational overhead introduced by the Wavelet decomposition and embedding modules is negligible relative to the cost of diffusion-based inference.

\begin{figure*}[t!]
\centering
\includegraphics[width=\linewidth]{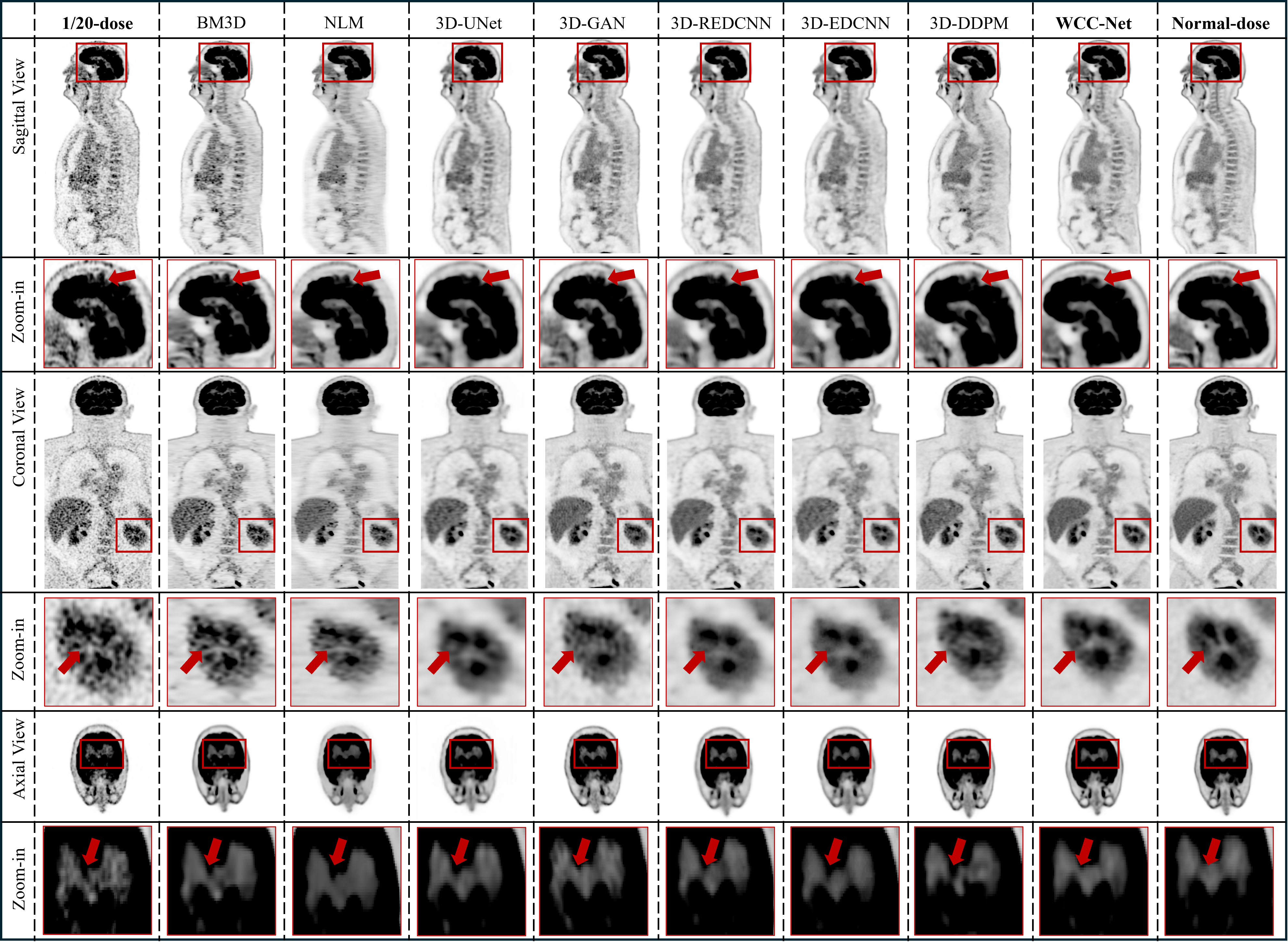}
\caption{Qualitative comparison of whole-body PET denoising results. Sagittal, coronal, and axial views are shown for a representative test subject, comparing the 1/20-dose input, competing denoising methods, and the normal-dose reference. Zoom-in regions (red boxes) highlight areas of high uptake and fine anatomical structures.}
\label{fig:comparison_figure}
\end{figure*}    

\section{Experiments}
\subsection{Datasets}\label{subsec:datasets}
We evaluate our method on the publicly available Ultra-Low-Dose PET (UDPET) Challenge dataset~\cite{xue2025udpet}
, which comprises whole-body PET scans acquired from multiple vendors and dose levels. To ensure consistency and minimize inter-scanner variability, we exclusively use 377 $^{18}$F-FDG PET scans acquired on the Siemens Biograph Vision Quadra system. Each subject provides paired normal-dose and ultra-low-dose PET images. The $1/20$-dose setting is used for supervised training and in-distribution evaluation, while additional dose levels ($1/4$ and $1/50$) are used to assess robustness under varying noise conditions.

The dataset is split at the subject level to prevent data leakage across experiments. In total, $297$ subjects are used for training, $20$ subjects for validation, and $60$ subjects for testing. The training set is used for both stages of our framework. The validation set is used exclusively for model selection and hyperparameter tuning, while the test set is reserved for final performance evaluation.

All PET images are converted to Standardized Uptake Value (SUV) units following the official preprocessing guidelines. Volumes are center-cropped to $192 \times 288$ in the axial plane and $520$ slices along the longitudinal axis, resulting in a final volume size of $192 \times 288 \times 520$. To facilitate efficient training of the 3D models and increase data diversity, each volume is further partitioned into overlapping 3D patches of size $96 \times 96 \times 96$, following the patch extraction strategy proposed by Yu \emph{et al.}~\cite{yu2025robust}. We adopt fully 3D modeling, as prior work has shown that 3D architectures outperform 2D approaches for PET denoising by leveraging volumetric context and structural continuity~\cite{yu2025robust}.

\subsection{Evaluation Metrics}\label{subsec:metrics}
For quantitative evaluation, we employ the Peak Signal-to-Noise Ratio (PSNR) and Structural Similarity Index Measure (SSIM)~\cite{psnrssim}, using the normal-dose PET images as the gold-standard. PSNR assesses voxel-wise reconstruction fidelity, while SSIM evaluates both structural similarity and perceptual image quality. To further evaluate structural preservation, we report the Gradient Magnitude Similarity Deviation (GMSD)~\cite{gmsd}, which compares gradient magnitude maps and is particularly sensitive to edge and texture distortions. In addition, we include the Normalized Mean Absolute Error (NMAE)~\cite{psnrssim} to quantify voxel-wise intensity deviations. This metric is especially relevant for PET imaging, as all volumes are converted to SUV units, making accurate intensity recovery critical for quantitative analysis. For statistical analysis, paired two-sided Wilcoxon signed-rank tests are conducted on per-case metrics between WCC-Net and each competing method. Holm correction is applied across the four metrics within each paired comparison, with significance defined at Holm-adjusted $p<0.05$ and $p<0.01$.
% ...
\begin{figure*}[t!]
\centering
\includegraphics[width=\linewidth]{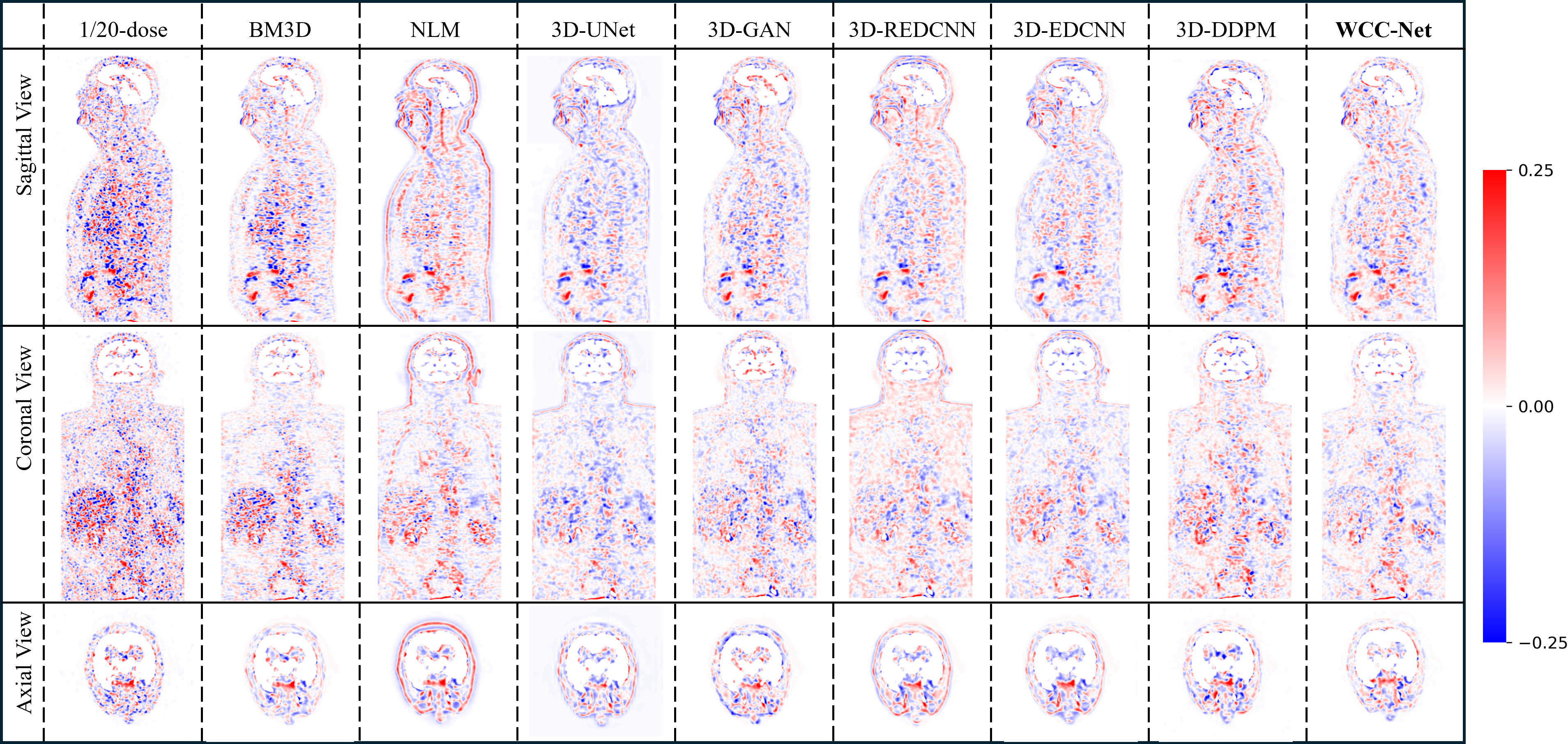}
\caption{Qualitative comparison of signed error maps for whole-body PET denoising at the 1/20 dose level. Sagittal, coronal, and axial views are shown for a representative test subject. Error maps are computed as the voxel-wise difference between each method and the normal-dose reference, and displayed using a diverging color scale centered at zero, where red and blue indicate over- and under-estimation, respectively.}
\label{fig:comparison_error_figure}
\end{figure*} 
\subsection{Implementation Details}\label{subsec:implementation}

All experiments are implemented using PyTorch and conducted on NVIDIA A100 GPUs. To ensure a fair and reproducible evaluation, we adopt the same training protocol across all compared methods. For diffusion-based models, the total number of diffusion timesteps is set to $T = 1000$, with a linear noise variance schedule where $\beta_t$ increases linearly from $1 \times 10^{-4}$ to $2 \times 10^{-2}$. 

All compared methods, including the diffusion backbone and the proposed WCC-Net, are trained using the Adam optimizer with an initial learning rate of $1 \times 10^{-4}$. For fair comparison on volumetric PET data, all baseline methods are implemented and evaluated in a fully 3D manner. Methods originally proposed in 2D~\cite{wang20183D-cGAN,redcnn,edcnn} are extended to 3D by replacing 2D operations with their 3D counterparts while preserving the original network design and training settings.

\section{Results}\label{sec:results}
  
\subsection{Quantitative Results}\label{subsec:quantitative_results}
We quantitatively compare WCC-Net with state-of-the-art PET denoising methods under both internal and external settings, covering seen and unseen dose levels. Performance is assessed using four complementary metrics: PSNR and SSIM (higher is better), and GMSD and NMAE (lower is better). 
A summary of all quantitative results is reported in Table~\ref{tab:quant_results}.

\subsubsection{Internal Evaluation:}\label{subsubsec:internal_evaluation} 
On the internal test set at the $1/20$ dose level, WCC-Net achieves the best performance across all metrics (Table~\ref{tab:quant_results}), with a PSNR of $43.594 \pm 1.404$ dB and an SSIM of $0.984 \pm 0.005$, outperforming both classical model-based approaches (BM3D~\cite{bm3d}, NLM~\cite{nlm}) and recent deep learning methods, including GAN-based~\cite{wang20183D-cGAN}, CNN-based~\cite{3dunet,redcnn,edcnn}, and DMs-based models~\cite{yu2025robust}. WCC-Net also yields the lowest distortion errors, with a GMSD of $0.011 \pm 0.003$ and an NMAE of $0.111 \pm 0.014$, demonstrating improved structural fidelity and reduced absolute error. 

Beyond absolute performance, we further analyze the relative improvements over the strongest competing baseline, namely, 3D DDPM~\cite{yu2025robust}. WCC-Net improves PSNR by $+1.21$ dB ($43.59$ vs. $42.38$ dB) and SSIM by $+0.008$ ($0.984$ vs. $0.976$), while further reducing both GMSD by $-0.003$ ($0.011$ vs. $0.014$) and NMAE by $-0.006$ ($0.111$ vs. $0.117$). These gains over competing methods are statistically significant (p < $0.05$ or p < $0.01$), illustrating the denoising effectiveness of the proposed method under matched training and testing conditions.
\begin{figure*}[t!]
\centering
\includegraphics[width=0.95\linewidth]{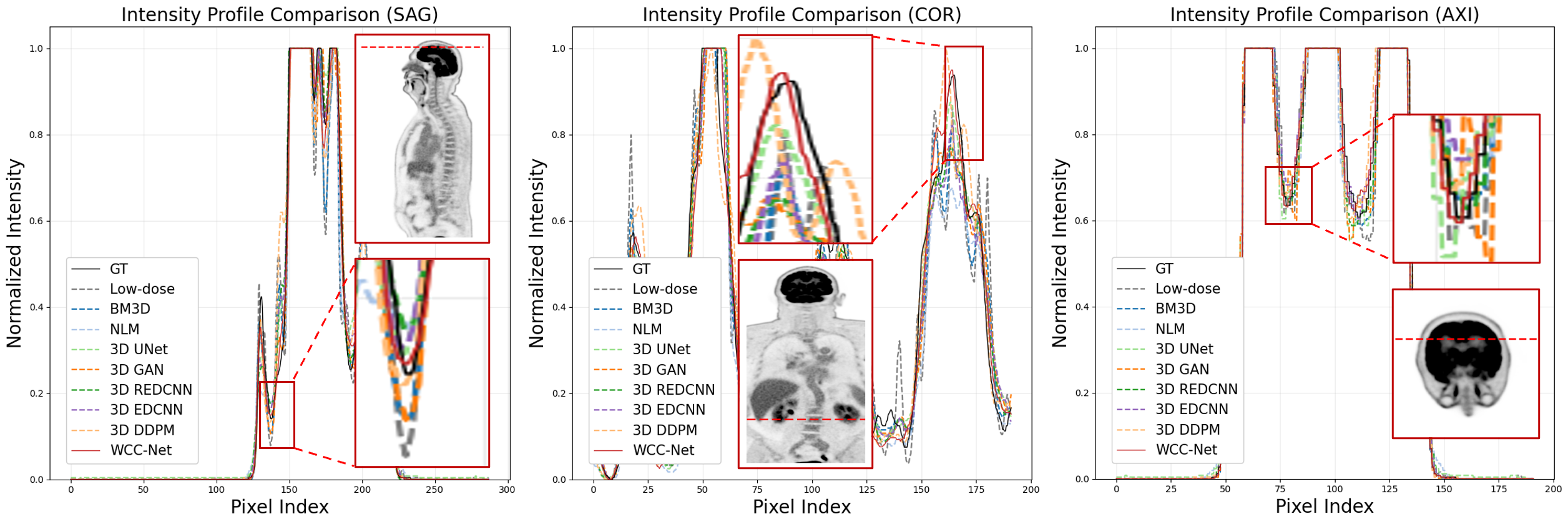}
\caption{Intensity profile comparison along representative sagittal, coronal, and axial directions for the same test subject at the 1/20 dose level. Line profiles are extracted along the red dashed lines indicated in the corresponding PET slices. Zoom-in regions highlight local differences in peak preservation and structural transitions.}

\label{fig:intensity_profile}
\end{figure*} 
\subsubsection{External Evaluation:}\label{subsubsec:external_evaluation} 
We further assess generalization performance on two external datasets acquired at unseen dose levels of $1/50$ and $1/4$.

At the external severely noisy $1/50$ dose level, WCC-Net continue to achieve superior performance, with a PSNR of $40.754 \pm 1.803$~dB and an SSIM of $0.976 \pm 0.011$ (Table~\ref{tab:quant_results}). At this unseen dose level, these improvements correspond to a PSNR gain of $+1.24$ dB and a SSIM improvement of $+0.012$ relative to the strongest baseline, 3D DDPM~\cite{yu2025robust}. Furthermore, WCC-Net achieves the lowest distortion errors, with GMSD of $0.014 \pm 0.006$ and NMAE of $0.132 \pm 0.027$, corresponding to absolute reductions of $-0.005$ ($0.014$ vs. $0.019$) and $-0.019$ ($0.132$ vs. $0.151$), respectively, compared to 3D DDPM~\cite{yu2025robust}.

At the moderately noisy $1/4$ dose level (Table~\ref{tab:quant_results}), several baselines benefit from the increased signal-to-noise ratio, as expected. 
WCC-Net consistently obtains the highest SSIM ($0.992 \pm 0.003$) and the lowest GMSD ($0.007 \pm 0.003$) among all methods. Its PSNR ($45.235 \pm 1.351$ dB) and NMAE ($0.084 \pm 0.013$) further confirm that WCC-Net helps to preserve fine-scale structural details, while minimizing the reconstruction error. Overall, WCC-Net generalizes more effectively than GAN-based~\cite{wang20183D-cGAN}, CNN-based~\cite{3dunet,redcnn,edcnn}, and DMs-based~\cite{yu2025robust} methods at unseen dose levels.

\subsection{Qualitative Results}\label{subsec:qualitative_results}

Fig.~\ref{fig:comparison_figure} shows visual comparisons of whole-body PET denoising results in sagittal, coronal, and axial views. Classical methods such as BM3D and NLM reduce noise but suffer from oversmoothing, leading to blurred organ boundaries and attenuation of small high-uptake structures. CNN- and GAN-based methods (3D-UNet~\cite{3dunet}, 3D-REDCNN~\cite{redcnn}, 3D-EDCNN~\cite{edcnn}, and 3D-cGAN~\cite{wang20183D-cGAN}) produce sharper images but often exhibit locally inconsistent textures or residual artifacts, particularly in regions with complex uptake patterns. Although the diffusion baseline yields globally coherent denoising, fine anatomical details, including thin cortical boundaries and small lesions, remain partially smoothed. In contrast, WCC-Net consistently preserves fine anatomical structures while effectively suppressing noise, producing images that more closely resemble the normal-dose reference, especially in magnified regions highlighting high-uptake organs and tissue interfaces.

To further assess reconstruction fidelity, Fig.~\ref{fig:comparison_error_figure} presents signed error maps. Competing methods exhibit spatially clustered over- and under-estimation, particularly near organ boundaries and regions of rapid intensity change, whereas WCC-Net shows lower error magnitude and reduced spatial bias. This advantage is further supported by the intensity profile analysis in Fig.~\ref{fig:intensity_profile}, where WCC-Net more closely follows the normal-dose reference across all anatomical directions, preserving peak uptake values and sharp transitions. These results demonstrate that wavelet-conditioned structural priors effectively guide the diffusion process toward anatomically faithful PET denoising.

\subsection{Ablation Studies} \label{subsec:ablation_studies}

\begin{table}[t!]
\centering
\caption{Ablation study on wavelet frequency selection.
LLL and HHH denote the lowest- and highest-frequency subbands, respectively.
All-High averages all subbands containing high-frequency components, All bands averages all eight subbands, and All-Low excludes the highest-frequency subband.}
\label{tab:ablation_table}
\setlength{\tabcolsep}{1pt}
\begin{tabular}{lccccc}
\toprule
\textbf{Setting} & \textbf{Freq.} 
& PSNR $\uparrow$ 
& SSIM $\uparrow$ 
& GMSD $\downarrow$ 
& NMAE $\downarrow$ \\
\midrule
Baseline & --
& 42.380  
& 0.976 
& 0.014  
& 0.117  \\
\midrule
A1 & LLL   
& \textbf{43.594 } 
& \textbf{0.984 }
& \textbf{0.011 }
& \textbf{0.111 } \\
A2 & HHH
& 42.817  
& 0.977 
& 0.013  
& 0.119  \\
A3 & All-High  
& 43.216  
& 0.982 
& 0.012  
& 0.117  \\
A4 & All-Low 
& 43.501  
& 0.983 
& 0.012  
& 0.112    \\
A5 & All bands 
& 43.485  
& 0.983  
& 0.013 5 
& 0.113  \\
\bottomrule
\end{tabular}
\end{table}
Table~\ref{tab:ablation_table} analyzes the effects of wavelet frequency selection within the proposed WCC-Net framework. Unless otherwise stated, wavelet-based structural priors are injected at the encoder stages of the diffusion backbone, which aligns with the role of wavelet representations as early, coarse structural guidance.

We first examine the impact of different wavelet frequency components. Injecting low-frequency wavelet coefficients (LLL, A1) yields the best overall performance across all metrics, confirming that low-frequency components provide a stable and noise-robust structural prior for PET denoising. In contrast, conditioning on the highest-frequency subband alone (HHH, A3) results in only marginal improvements over the baseline, indicating that noise-dominated high-frequency components are less suitable as effective structural guidance. Averaging high-frequency subbands (All-High, A4) improves performance relative to HHH but remains inferior to LLL-based conditioning, suggesting that high-frequency information contains limited complementary structural cues. Excluding the highest-frequency subband (All-Low, A5) further improves performance and approaches LLL-based results, indicating that suppressing extreme high-frequency noise leads to more reliable guidance. Incorporating all wavelet subbands jointly (All bands, A6) improves over the baseline but does not surpass LLL-only conditioning, implying that high-frequency components may dilute the benefits of low-frequency structural priors. Overall, encoder-side conditioning with low-frequency wavelet components provides the most effective and consistent gains.

\section{Conclusion and Future Works}

In this work, we proposed WCC-Net, a wavelet-conditioned diffusion framework for whole-body low-dose PET denoising. By injecting wavelet-based structural priors into a frozen diffusion backbone via ControlNet-style conditioning, WCC-Net effectively decouples anatomical structure from noise and promotes anatomically consistent denoising. Extensive experiments on ultra-low-dose PET demonstrate that WCC-Net consistently outperforms CNN-, GAN-, and DMs-based baselines across multiple dose levels, achieving improved quantitative performance and better preservation of diagnostically relevant structures. These results highlight the effectiveness of frequency-domain structural priors for controlled diffusion under low signal-to-noise conditions.

Despite these promising results, limitations remain. Our current implementation employs a fixed single-level Haar wavelet decomposition, which may restrict modeling of multi-scale anatomical structures. Moreover, evaluation is limited to a single dataset acquired on one scanner using a single radiotracer ($^{18}$F-FDG), which may affect generalizability.

Future work will investigate more flexible wavelet representations, including alternative wavelet bases and multi-level decompositions, and extend validation to multi-vendor and multi-site datasets with diverse tracers and acquisition protocols. Incorporating clinical validation, such as expert assessment and task-driven evaluation, will further facilitate translation to real-world PET imaging.

\section*{Acknowledgements}
Javier Montoya is supported by the Swiss National Science Foundation (SNSF) under grant number 20HW-1 220785. Guang Yang was supported in part by the ERC IMI (101005122), the H2020 (952172), the MRC (MC/PC/21013), the Royal Society (IEC/NSFC/211235), the NVIDIA Academic Hardware Grant Program, the SABER project supported by Boehringer Ingelheim Ltd, NIHR Imperial Biomedical Research Centre (RDA01), The Wellcome Leap Dynamic resilience program (co-funded by Temasek Trust)., UKRI guarantee funding for Horizon Europe MSCA Postdoctoral Fellowships (EP/Z002206/1), UKRI MRC Research Grant, TFS Research Grants (MR/U506710/1), Swiss National Science Foundation (Grant No. 220785), and the UKRI Future Leaders Fellowship (MR/V023799/1, UKRI2738). Peiyuan Jing is supported by the Swiss National Science Foundation (SNSF) under grant number 20HW-1 220785.
\nocite{langley00}

\bibliography{example_paper}
\bibliographystyle{icml2025}



\end{document}